\documentclass[letterpaper, 10 pt, conference]{ieeeconf}  % Comment this line out if you need a4paper

\usepackage{graphicx}
\usepackage{threeparttable}
\usepackage{subfig}
\usepackage{algorithmicx}
\usepackage[ruled]{algorithm2e}
\IEEEoverridecommandlockouts                              % This command is only needed if 
\title{\LARGE \bf
Disentangling and Vectorization: A 3D Visual Perception Approach for Autonomous Driving Based on Surround-View Fisheye Cameras
}

\author{
 Zizhang Wu$^{1}$  \qquad
 Wenkai Zhang$^{1}$  \qquad 
 Jizheng Wang$^{1}$  \qquad 
 Man Wang$^{1}$    \qquad \\
 Yuanzhu Gan$^{1}$  \qquad
 Xinchao Gou$^{2}$  \qquad
 Muqing Fang$^{1}$  \qquad
 Jing Song$^{1}$% <-this % stops a space
\thanks{$^{1}$Zongmu Technology
        }%
\thanks{$^{2}$Karlsruhe Institute of Technology
        }%
}

\begin{document}

    %%%%%%%%%%%%%%%%%%%%%%%%%%%%%%%%%%%%%%%%%%%%%%%%%%%%%%%%%%%%%%%%%%%%%%%%%%%%%%%%
    \maketitle
    \thispagestyle{empty}
    \pagestyle{empty}
    %%%%%%%%%%%%%%%%%%%%%%%%%%%%%%%%%%%%%%%%%%%%%%%%%%%%%%%%%%%%%%%%%%%%%%%%%%%%%%%%
    \begin{abstract}
The 3D visual perception for vehicles with the surround-view fisheye camera system is a critical and challenging task for low-cost urban autonomous driving. While existing monocular 3D object detection methods perform not well enough on the fisheye images for mass production, partly due to the lack of 3D datasets of such images. In this paper, we manage to overcome and avoid the difficulty of acquiring the large scale of accurate 3D labeled truth data, by breaking down the 3D object detection task into some sub-tasks, such as vehicle's contact point detection, type classification, re-identification and unit assembling, etc. Particularly, we propose the concept of Multidimensional Vector to include the utilizable information generated in different dimensions and stages, instead of the descriptive approach for the bird's eye view (BEV) or a cube of eight points. The experiments of real fisheye images demonstrate that our solution achieves state-of-the-art accuracy while being real-time in practice.

\end{abstract}
    %%%%%%%%%%%%%%%%%%%%%%%%%%%%%%%%%%%%%%%%%%%%%%%%%%%%%%%%%%%%%%%%%%%%%%%%%%%%%%%%
    \section{INTRODUCTION}

In recent years, autonomous driving has attracted more and more attention from both industry and research communities \cite{2020IROSVision}. The 3D perception of the environment is one of the most challenging tasks \cite{2020IROSAccurate}.
Recent advances \cite{2016arxivvehicle} demonstrate the potential to replace the LIDAR with cheap onboard cameras, which are readily available on most modern vehicles.
Particularly, the surround-view fisheye camera system is very popular in mass production, on account of a larger field-of-view (FoV) \cite{2018IVreal} than the pin-hole cameras. 

The surround-view fisheye camera system can provide a 360-degree perception, which makes up for other pinhole cameras' lack of close-range perception around the vehicle, especially in the scene of a traffic jam. So more researches focus on obtaining position and pose information from the above system, which are mostly in the Bird's Eye View (BEV) manner \cite{ng2020bev}.
%The estimation of the positions and heading angles of surrounding vehicles is fundamental to trajectory prediction and planning.
%Particularly, the powerful Bird-Eye View (BEV) method has gain popularity for representing the environment in a suitable manner \cite{ng2020bev}. 
To obtain a BEV image, previous works either use the Inverse Perspective Mapping (IPM) method \cite{mallot1991inverse} to stitch the raw image data from the surrounding cameras or leverage 3D object detection methods \cite{2016arxivvehicle,2019ICCVaccurate, 2019AAAImonogrnet} to obtain the 3D bounding boxes of the target-vehicles, which are then projected on the BEV image. 
Despite its simplicity, the IPM method can not directly provide valuable information such as the positions and the heading angles of the target-vehicles \cite{2019IVdeep}. 
On the contrary, 3D object detection methods \cite{2019CVPRmulti,2020CVPRWsmoke} can provide more comprehensive information, but a fisheye image dataset with labeled 3D truth data is hardly available \cite{2019ICCVwoodscape}.
Annotating such datasets is costly and heavily relies on the fisheye cameras used to collect data, which makes it unsuitable for mass production.
Meanwhile, 3D object detection requires more computing resources compared to 2D \cite{2016cvpryou}.
Existing approaches usually utilize the visual features to regress the heading angles of the target-vehicles \cite{headingangle,headingangle1} and achieve great success in many public datasets such as KITTI \cite{KITTI}.
However, it remains a challenging task to obtain robust results covering a wide variety of use cases in industrial applications \cite{2019arxivfisheyemultinet}.

\begin{figure}[!t]
    \centering
    \includegraphics[width=9cm]{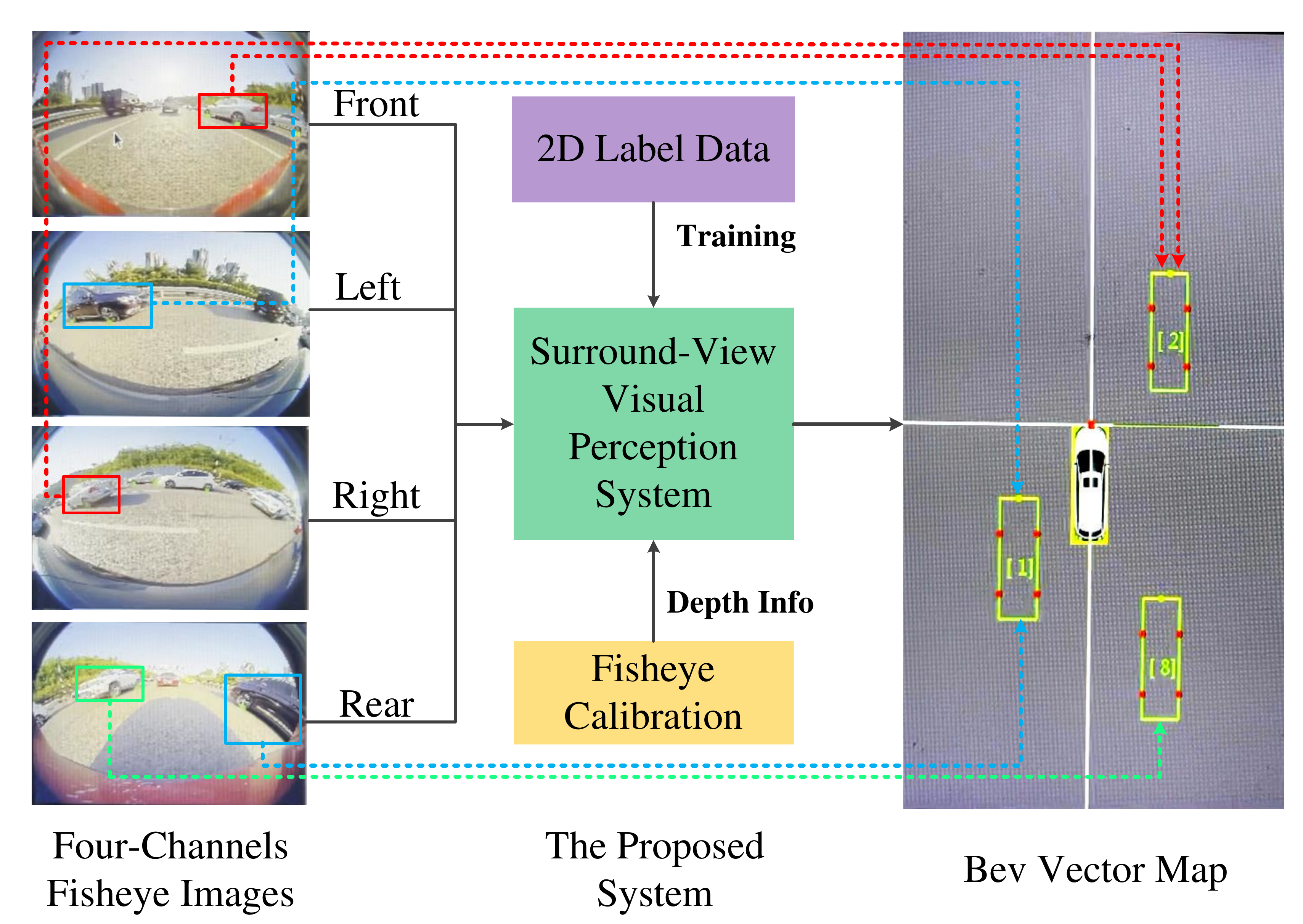}
    \caption{The overview of the proposed method. Best view in color and zoom in.}
    \label{fig1}
\end{figure}

\begin{figure*}[!t]
    \centering
    \includegraphics[width=18cm]{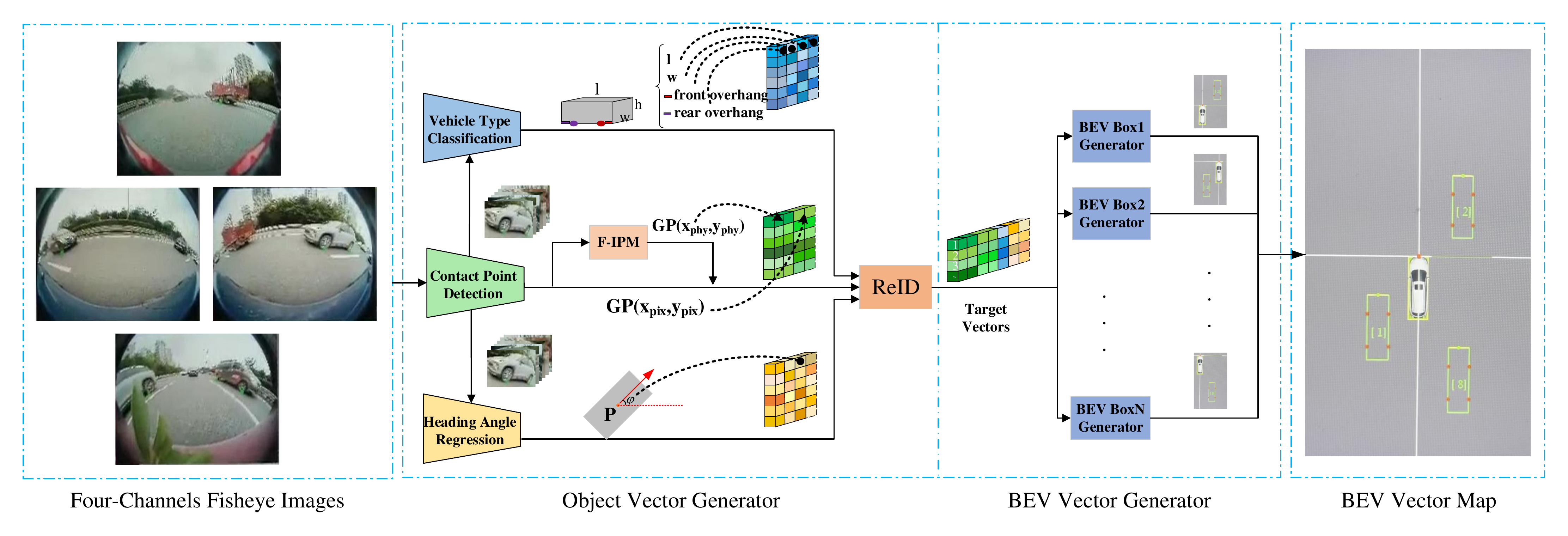}
    \caption{The overall structure of our proposed approach. The inputs are four-channels fisheye images which construct a surround-view environment for ego-vehicles. The final output is a vector map containing the shape of the object under the bird's-eye view. Best view in color and zoom in.}
    \label{fig_main_framework}
\end{figure*}

% Occupying the niche

In autonomous driving and mobile robotics applications, we can generally assume that the most important dynamic objects are on a ground plane, and the camera is mounted at a certain height above the ground \cite{2021IRALground}. In particular, the assumptions hold well in quite close distance, such as within the perception range of the surround-view fisheye camera system \cite{2019IVdeep}.
So we can obtain the 3D positions of certain points on the ground through inverse projection, and these points can serve as useful hints under the condition of being stable and detectable in the image.

In the past few years, the two-dimensional keypoint detection and object detection techniques are gradually mature \cite{2016cvpryou,2019arxivobjects}. Inspired by the above two sides, we define some keypoints of the vehicle in the image, which are also the contact points on the ground \cite{2021IRALground}. 
As mentioned above, we can acquire the 3D positions of the vehicle's contact points, such as the contact points of wheels and bumpers. But we can't compute the three-dimensional positions only based on the above information. For example, in some situations, we also need the heading angle and vehicle type for additional information, which is bounded with its types such as the length, width and height. So we disentangle the 3D object detection task into some sub-tasks, such as vehicle's contact point detection, type recognition, and heading angle regression. Moreover, to maximize the advantages of every algorithm component, we establish a robust mechanism to achieve stable acquisition of 3D information for different situations, combined with geometric constraints. Furthermore, we propose the Multidimensional Vector to improve the prediction effect, which describes the target-vehicle by its center point, heading angle, type, etc. So in this paper, we propose a system that takes raw images from four surrounding fisheye cameras as input and outputs the 3D position of the target-vehicle with only the 2D labeled truth data (including some heading angle truth data) and calibration required, as shown in Fig. \ref{fig1}.

The contributions are summarized as following three-fold:
1. We provide a solution for 3D visual perception of the surrounding environment with a fisheye camera system in autonomous driving and avoid the difficulty of acquiring a large scale of accurate 3D labeled truth data.
2. We improve the prediction effect by the proposed Multidimensional Vector, which includes the utilizable information generated in different dimensions and stages, for more robust re-identification and tracking, etc.
3. Our method demands lower computing resources and achieves state-of-the-art accuracy in real-time scenes, thus has great advantages in mass production.

    %%%%%%%%%%%%%%%%%%%%%%%%%%%%%%%%%%%%%%%%%%%%%%%%%%
\section{Related Works}

% %---------------------

% In this section, we will first review recent advances in surround-view fisheye cameras and 3D visual perception for autonomous driving with the surround-view fisheye camera. Then we will discuss the representation of the usual 3D detection. 

\subsection{Surround-view fisheye cameras and 3D visual perception}

Extensive visual perception tasks usually employ surround-view fisheye cameras to capture the surrounding environment due to their large FoV \cite{2018IVreal} and sufficiently good performance \cite{2018IROSend,2014ICCPomnidirectional}. In particular, ego-vehicles can achieve 360-degree perception using only four fisheye cameras, which is conducive to mass production \cite{2020ICRAfisheyedistancenet}. \cite{2015IV360} proposes a system to classify track vehicles and pedestrians around the ego-vehicle using only four fisheye cameras. \cite{2018IVreal} presents a solution to detect and classify moving objects around the vehicle in real-time, which merges four views captured by fisheye cameras into a single frame. Moreover, surround-view fisheye cameras are widely used in various perception tasks, such as semantic segmentation \cite{ 2018ECCVWsemantic}, depth estimation \cite{2018CVPRWnear, 2021WACVsyndistnet}, object detection \cite{2020accessfisheyedet,2019arxivfisheyemodnet} and Re-Identification (ReID) \cite{2020CVPRWvehicle}.

%---------------------------------------------------

Despite the advantages mentioned above, fisheye cameras have a strong radial distortion and exhibited more complex projection geometry \cite{2019ICCVWrotinvmtl}, which leads to appearance distortion \cite{2020arxivmonocular}. Thus, it is hard to generalize the models trained on the fisheye dataset. 
Besides, the distortion brings additional challenges to annotate fisheye datasets \cite{2019ICCVwoodscape,2019ICCVWdesoiling}, and there are few public 3D visual perception systems for vehicles based on the surround-view fisheye cameras \cite{2019arxivfisheyemultinet}. Moreover, OmniDet \cite{2021arxivomnidet} presents a multi-task visual perception network to capture the surrounding environment. Their object detection task requires a large annotated dataset of fisheye images which increases the additional cost of the integral task.
The dense pixel-wise depth estimation would result in prohibitive computational cost and limited real-time performance.
%------------------------------------------------

\subsection{3D object detection}

The visual perception system is a crucial part of autonomous driving, in which 3D object detection is an important component to estimate the pose and location of the surrounding vehicles \cite{2016arxivvehicle, 2021IET4net}.
% 3D object detection is a more challenging task compared with the 2D task because of the additional dimensions \cite{2016cvpryou,2019arxivobjects}.
Recent developments \cite{2020CVPRWsmoke,2019AAAImonogrnet,2019CVPRmulti} in 3D object detection mainly utilize various information such as context, geometry, depth map, and so on. They usually regress 3D information of target-vehicles, including the bird's eye view or a cube of eight points, through neural networks \cite{2016cvprmonocular, 2017CVPRdeep,2019ICCVm3d}.

However, previous representations provide only limited information for the visual perception task in autonomous driving.
In this paper, we propose the Multidimensional Vector that includes more usable information, instead of the descriptions of the bird's eye view or a cube of eight points.

\begin{figure}[!b]
    \centering
    \includegraphics[width=8.5cm]{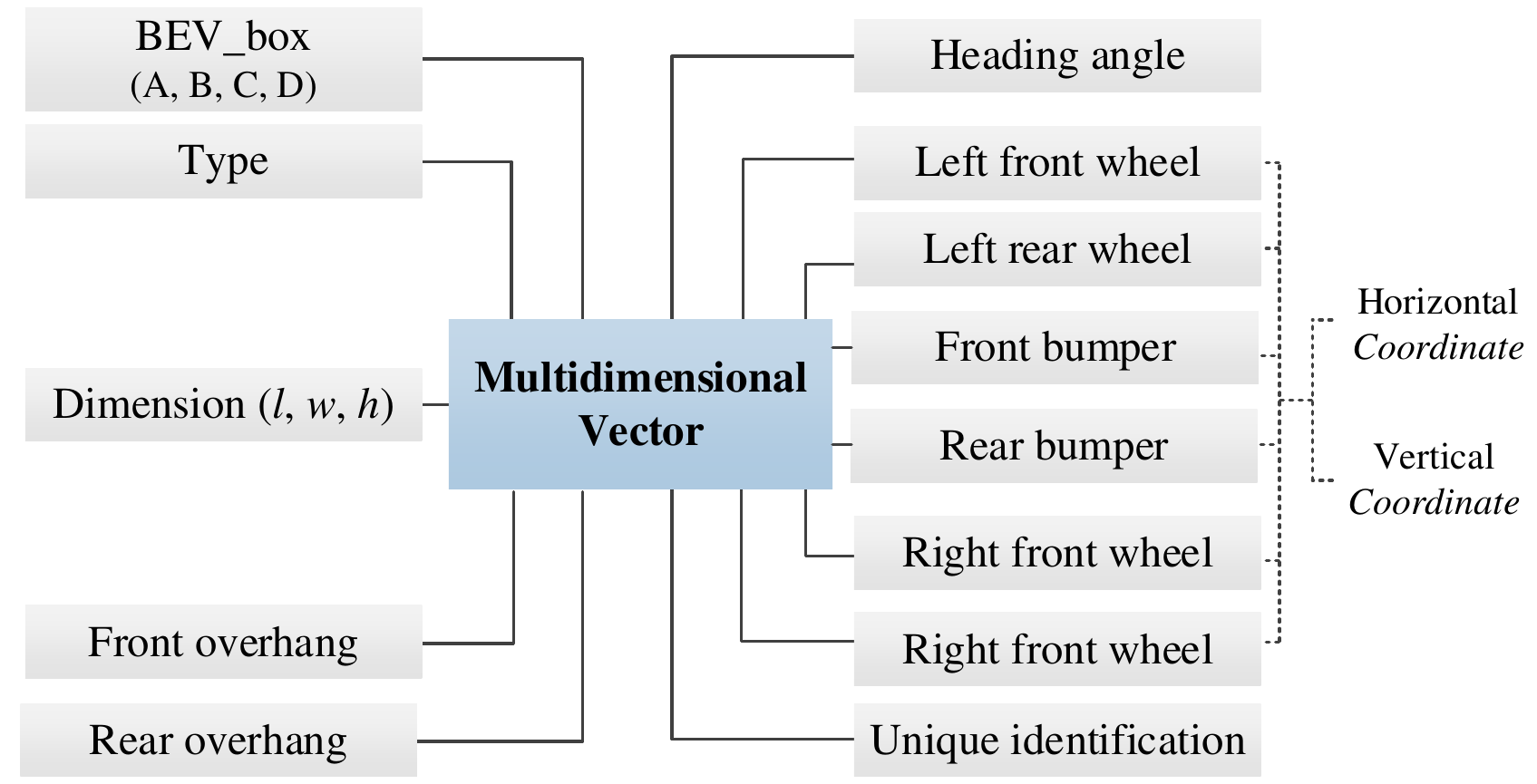}
    \caption{The specific composition of the Multidimensional Vector.}
    \label{fig_multivector}
\end{figure}

\section{Methods}
In this section, we introduce the proposed visual perception system in detail, as shown in Fig. \ref{fig_main_framework}. 
First of all, we feed four fisheye images into the Contact Point Detection module, Vehicle Type Recognition module, and Heading Angle Regression module to obtain the required attributes for Multidimensional Vector, including vehicle type, dimensions $(l, w, h)$, front overhang $(fo)$, rear overhang $(ro)$, heading angle $(\varphi)$ and contact points' coordinates, as shown in Fig. \ref{fig_multivector}. Then these attributes are merged and unified by the ReID module to generate a unique identification number for each target-vehicle. Subsequently, we integrate the intermediate results to generate vectors for the BEV description of target-vehicles and illustrate the final BEV Vector Map. Table \ref{tab:table1} shows the notations used in this paper.

\begin{table}[h!]
    \begin{center}
    \caption{Notations used in this paper.}
    \label{tab:table1}
        \begin{threeparttable}
            \begin{tabular}{l|l}
            % \toprule 
            \hline
              \textbf{Notations} & \textbf{Descriptions}\\
            %   $\alpha$ & $\beta$ & $\gamma$ \\
            %   \hline
            %   \multirow{2}{*}{12} & 1110.1 & a\\ % <-- Combining 2 rows with arbitrary with (*) and content 12
            %   & 10.1 & b\\ % <-- Content of first column omitted.
            %   \midrule 
              \hline
              $A,B,C,D$ & \begin{tabular}[c]{@{}l@{}}BEV\_box \\ (Left-Front, Left-Rear, Right-Front, Right-Rear) \end{tabular}   \\
              \hline
              $FW$ & Front wheel ground contact point \\
              \hline  
              $RW$ & Rear wheel ground contact point \\
              \hline
              $FB$ & Front bumper contact point \\
              \hline
              $RB$ & Rear bumper contact point \\
              \hline
              $P$ & Center point of the target vehicle  \\
              \hline
              $O$ & Origin point of the ego-vehicle coordinate system \\
              \hline
              $GP_{{phy}}$ & Physical coordinates of the ground contact point \\
              \hline
              $GP_{{pix}}$ & Pixel coordinates of the ground contact point \\
              \hline
              $l,w,h$ & length,width and height of the target vehicle\\
              \hline
              $fo$ & Front overhang of the target vehicle\\
              \hline
              $ro$ & Rear overhang of the target vehicle\\
              \hline
              $obj_{{id}}$ & Id of the target vehicle\\
              \hline
              $\theta\tnote{*}$ & Azimuth angle of the target-vehicle\\
              \hline
              $\varphi\tnote{*}$ & Heading angle of the target-vehicle\\
              \hline
              $\gamma\tnote{*}$ & Angle of $line(RB, RW)$ in standard position \\
              \hline
            %   \bottomrule 
            \end{tabular}
            \begin{tablenotes}
            \footnotesize
                \item[*] All the angles are positive angles along the positive x-axis of the ego-vehicle coordinate system.
            \end{tablenotes}
        \end{threeparttable}
    \end{center}
\end{table}

\subsection{Object Vector Generator}
\subsubsection{Contact Point Detection}
To overcome and avoid the difficulty of acquiring large-scale accurate 3D labeled truth data, we perform 2D contact point detection on fisheye images. For detecting and locating the surrounding vehicles, we need to detect vehicles, front wheels, and rear wheels to compute the contact points of target-vehicles. Considering the trade-off between real-time performance and accuracy in autonomous driving scenes, we employ the CenterNet \cite{2019arxivobjects} as the base framework. Specially, we input four fisheye images from different channels into the network, which outputs the predicted bounding boxes of the vehicles, front and rear wheels. 

\begin{figure}[!t]
    \centering
    \includegraphics[width=8cm]{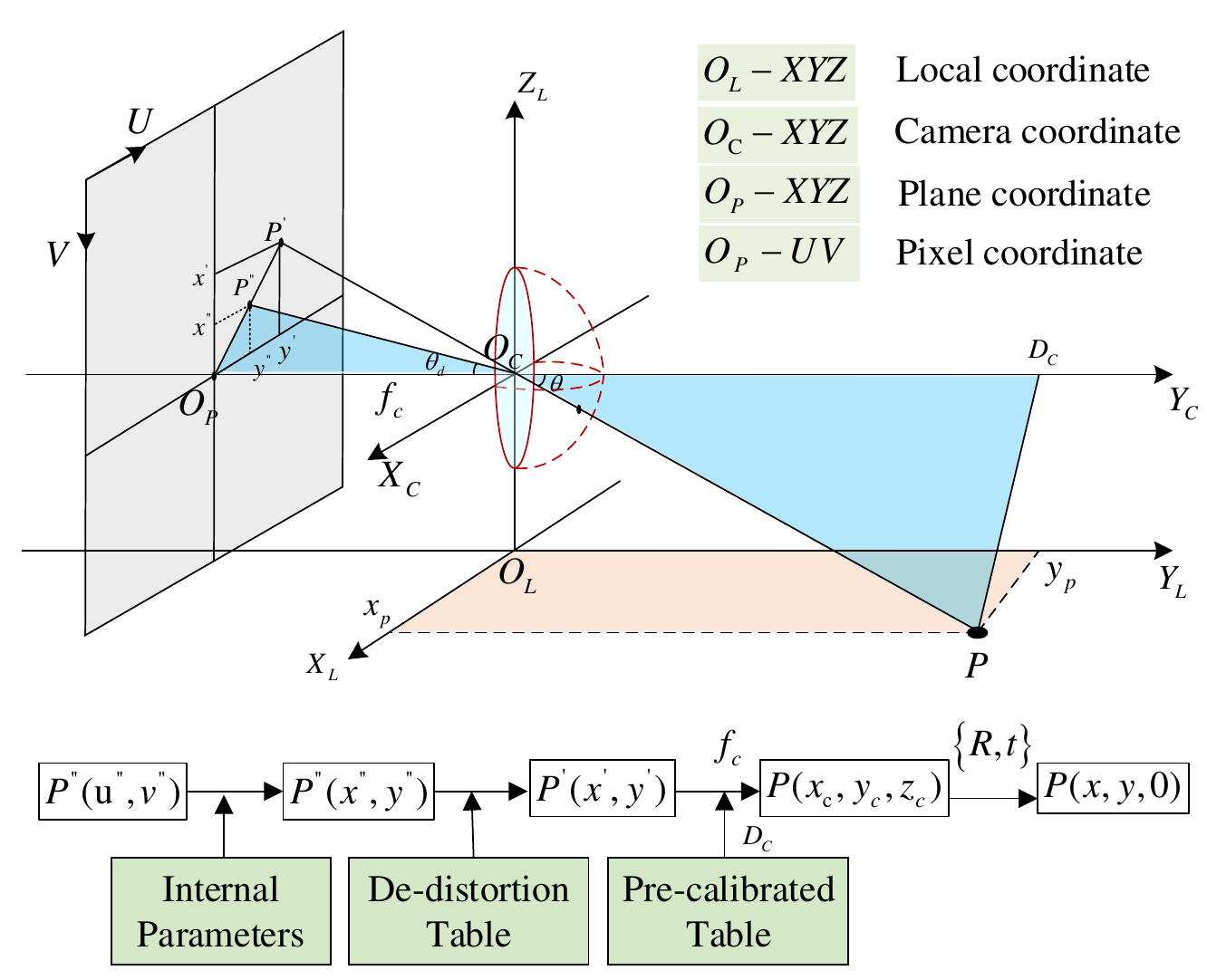}
    \caption{The diagram of translating the pixel coordinates of the contact points in fisheye image to the physical coordinates. Best view in color and zoom in.}
    \label{fig_IPM}
\end{figure}

Then we acquire the contact points' pixel coordinates in the image through post-processing. Here we define (ground) contact point as the midpoint of the bottom edge of the bounding box of a vehicle or a wheel. These points are on the ground and their physical coordinates are $P(x, y, 0)$. With the pixel coordinates of the contact points, we can calculate their physical coordinates in the real-world coordinate system through the Fisheye IPM algorithm \cite{mallot1991inverse}, as shown in Fig. \ref{fig_IPM}.
%------IPM-------------
The pixel coordinates of contact points in fisheye images are $P^{\prime\prime}(u^{\prime\prime},v^{\prime\prime})$, which are transformed into the distorted plane with coordinates $P^{\prime\prime}(x^{\prime\prime},y^{\prime\prime})$ through the internal parameter matrix.
By referring to the de-distortion table, they can be converted into the de-distorted plane with coordinates $P^{\prime}(x^{\prime},y^{\prime})$.
%P'(x', y'). 
After introducing the re-calibrated internal parameters and pre-calibrated tables, we get coordinates $P(x, y, z)$ in the camera coordinate system and target depth values $D_c$.
Through the $R, t$ matrix of the external parameters, we obtain coordinates $P(x, y, 0)$ of contact points in the local coordinate system (ego-vehicle coordinate system).
%-------------------
In conclusion, we prepare some useful information of detected vehicles in this step, including 2D bounding boxes, pixel coordinates ($GP_{pix}$) and physical coordinates ($GP_{phy}$) of the contact points ($FB$, $RB$, $FW$, $RW$).

\subsubsection{Vehicle Type Classification}
We infer more information on target-vehicles by our vehicle type classification module, which contains a simple classification network \cite{lecun1998gradient}. We crop vehicle images according to the detected 2D bounding boxes and input them into the network, which outputs the types of vehicles. Particularly, we divide vehicles into 8 types, including car, SUV, MPV, etc. Therefore, we obtain additional attributes for Multidimensional Vector, such as the dimensions $(l, w, h)$, front overhang $(fo)$, and rear overhang $(ro)$, because they are bonded with the type of vehicles.

% In order to obtain more vehicle information, we need to determine the type of vehicles. Our type classification module includes a simple classification network \cite{lecun1998gradient}. We input the cropped vehicle images according to the detected 2D bounding box into the network and the type of the vehicles will be output. The type of vehicles includes car, SUV, MPV and so on. According to these types, we can esaily obtain the dimension (l, w, h), front overhang and rear overhang which are attributes of the Multidimensional Vector.

\subsubsection{Heading Angle Regression}
The Heading Angle Regression module is implemented by the MultiBin \cite{headingangle}, which is a method to estimate the pose of objects by the cropped images. We take the output of this module as the estimation of the heading angle in the third case mentioned in \ref{3_cases}.
%(HAngle($\varphi$))

\subsubsection{ReID}
In our surround-view system, a target-vehicle can be observed from different cameras and described by one or more vectors. To facilitate the subsequent tasks, it is necessary to describe the target-vehicle using a vector with a unique identification number. So, we merge the original vectors from three branches into one vector with the unique identification number, as shown in Fig. \ref{fig5}. This process is target re-identification (ReID) and consists of three different stages.
\begin{figure}[!t]
    \centering
    \includegraphics[width=8cm]{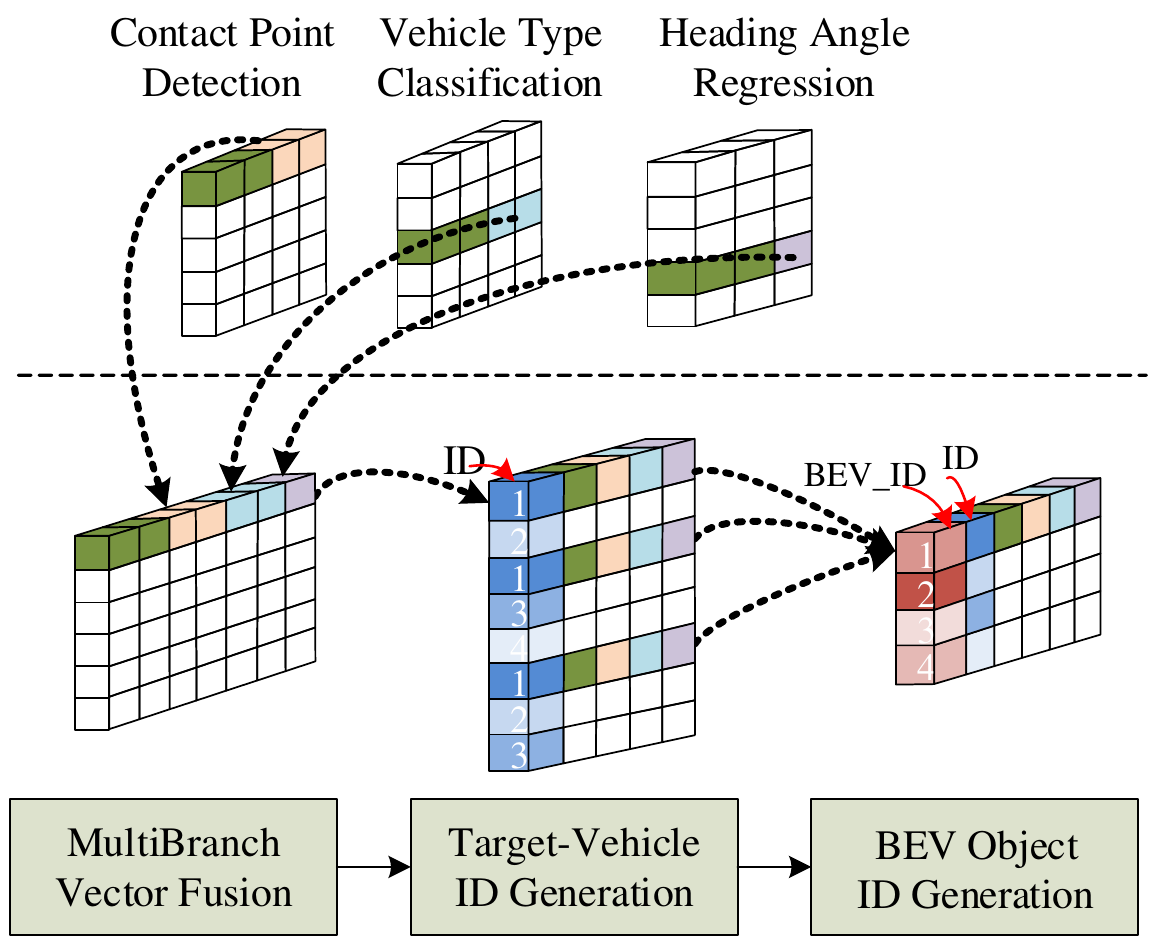}
    \caption{ReID module is divided into three stages. The first stage fuses the vectors of three branches, the scond stage generates an ID for each object of each channel, and the third stage merges the vectors which describing the same object into one vector. Best view in color and zoom in.}
    \label{fig5}
\end{figure}

The first stage is the multi-branch vector fusion. The bounding boxes $(X, Y, W, L)$ information of the target-vehicles is included in the vector of each branch. If two vectors in different branches have the same bounding box, we append one to the other and remove duplicate elements. In this instance, we append the vectors from the vehicle type classification and heading angle regression branches to the contact point detection vector branch.

The second stage is the target-vehicle id generation. Since the same target may be observed in different channels and described with different vectors, the goal of this stage is to assign the same identification number to these vectors. This stage is mainly achieved by some hand-craft rules. For example, we assign the same identification number to vectors, whose physical coordinates of ground contact points are less than 50 cm away.
 
The third stage is the BEV target id generation, which aims to merge all the vectors that describe the same target-vehicle and assign them a unique identity. Two situations are considered in this stage. One is the fusion between channels, which is performed in a complementary or weighted way, as shown in Fig. \ref{fig5}(1). The other is the fusion between categories, which is carried out in a complementary way, as shown in Fig. \ref{fig5}(2). Finally, we get the unique vector representation that contains all the information for the BEV Vector Map.
\begin{figure}[!t]
    \centering
    \includegraphics[width=8cm]{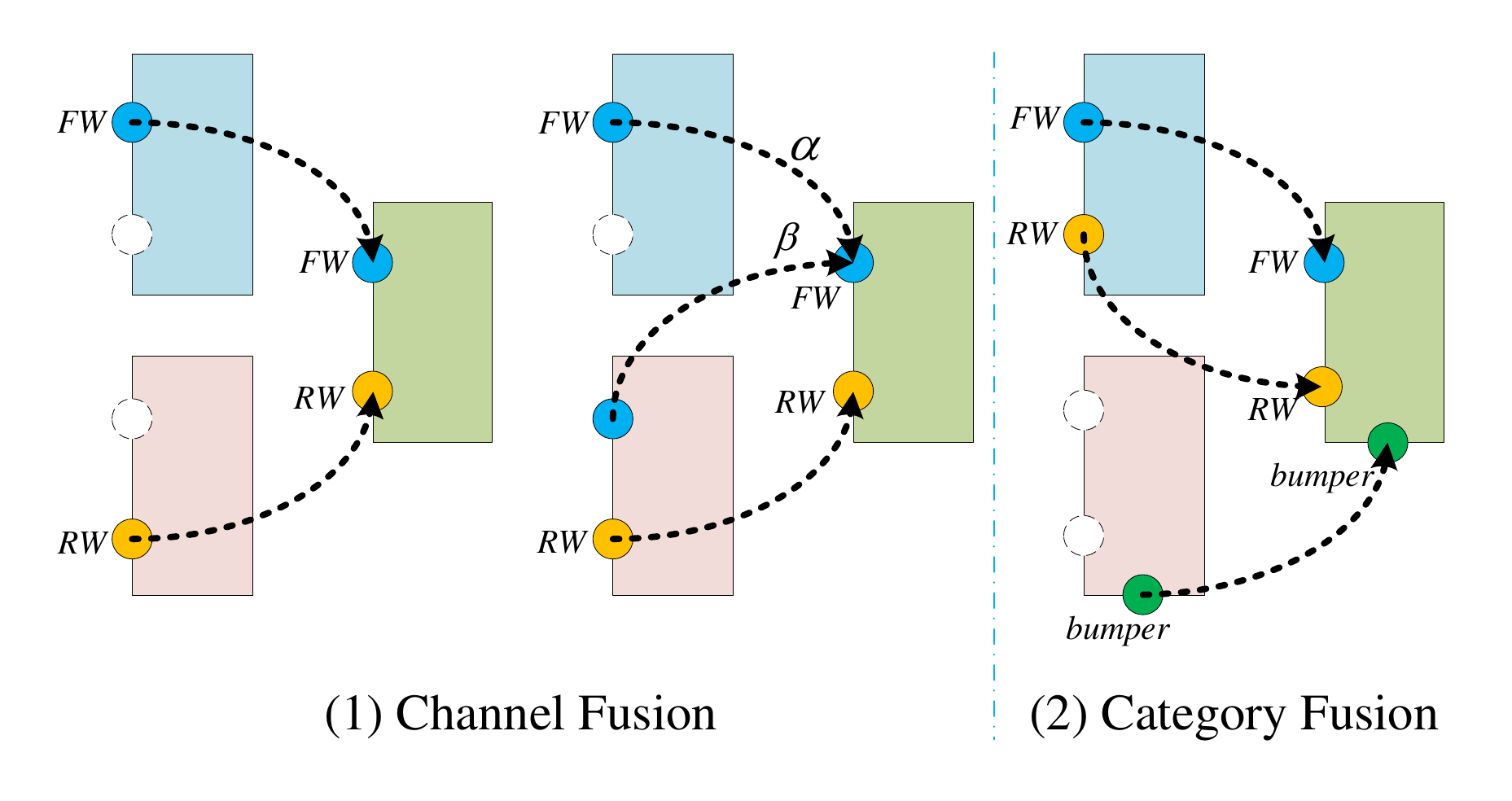}
    \caption{Channel fusion and category fusion. (1) Wheels from different channels are fused into one vehicle. $\alpha$ and $\beta$ are 0.5, which indicate weights assigned to the front wheels of two vehicles. (2) Wheels and bumper from different categories are fused into one vehicle. Best view in color and zoom in.}
    \label{fig5}
\end{figure}

\subsection{BEV Vector Generator} \label{3_cases}
The output of the previous modules is integrated to calculate the BEV Vectors, which are used to construct the BEV Vector Map.

Firstly, we estimate the position of the center point $P$ and the heading angle $\varphi$ of the target-vehicle.
The azimuth angle $\theta$ and the heading angle $\varphi$ (as shown in Table \ref{tab:table1}) are used to determine whether the visible side is left or right side of the target-vehicle via equation (\ref{equation_left_right_side}).
The azimuth angle can be calculated by the position of the wheels or the bumpers.
\begin{equation} \label{equation_left_right_side}
    \left\{
        \begin{array}{lr}
            sin(\varphi - \theta) > 0, & left \ side \\
            sin(\varphi - \theta) < 0, & right \ side
        \end{array}
    \right.
\end{equation}

Then, three different cases are considered to estimate the center point $P$ and the heading angle $\varphi$ of the target-vehicle.
\begin{figure}
  \begin{minipage}[t]{0.5\linewidth}
    \centering
    \includegraphics[scale=0.45]{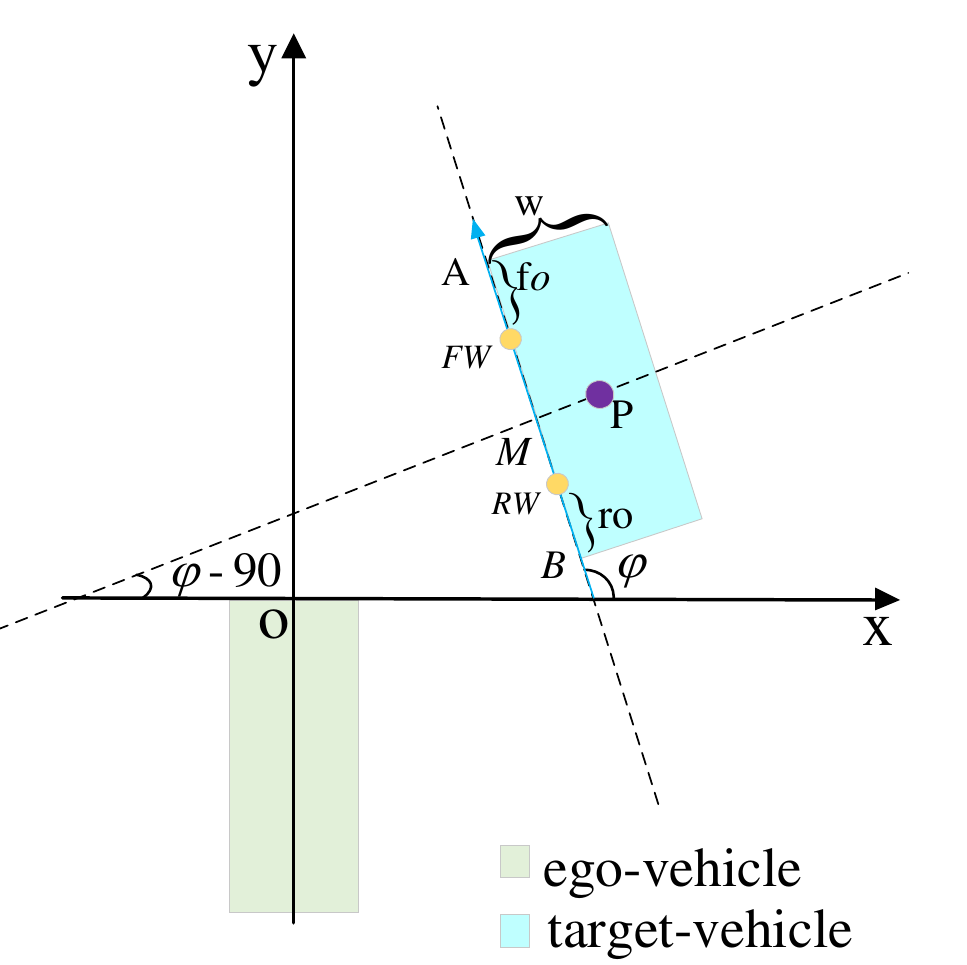}
    \centerline{(a)}
  \end{minipage}%
  \begin{minipage}[t]{0.5\linewidth}
    \centering
    \includegraphics[scale=0.45]{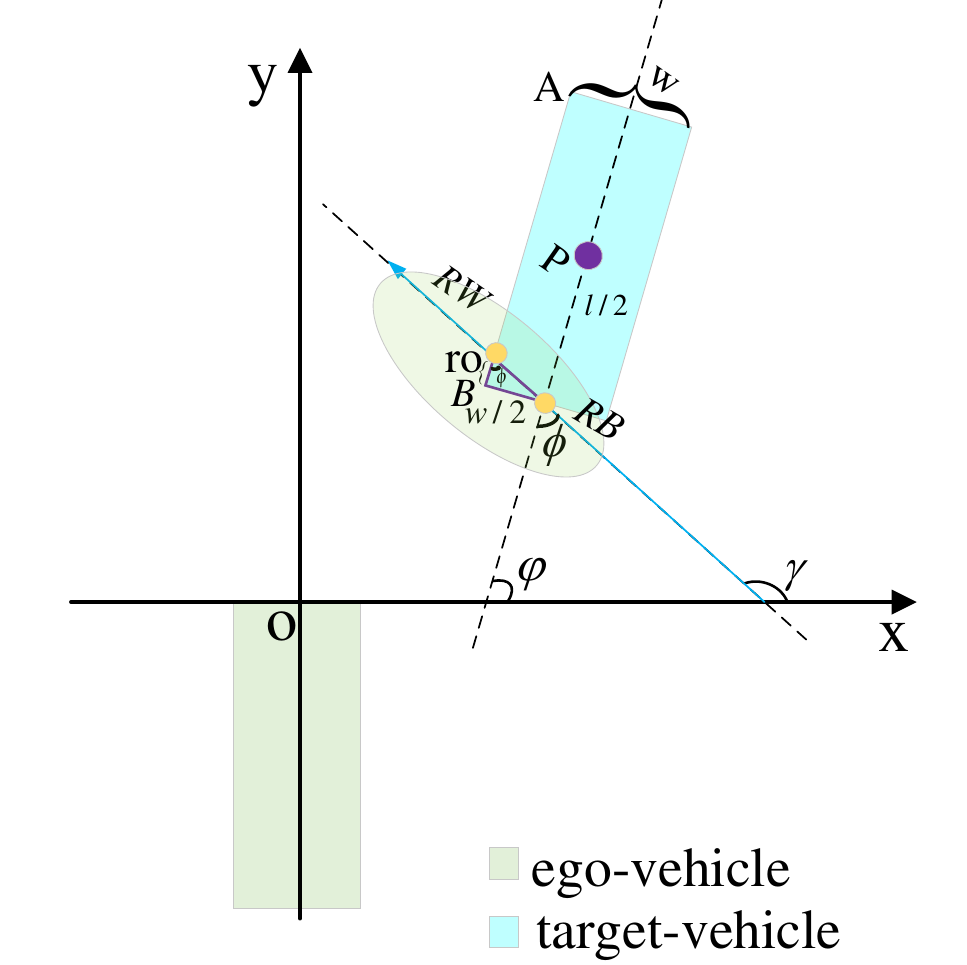}
    \centerline{(b)}
  \end{minipage}
\caption{Two cases of calculating the heading angles of the target-vehicles. (a) two wheels on one side are visible. (b) only one wheel and one bumper are visible. Best view in color and zoom in.}
\label{f_3.2.2}
\end{figure}
In the first case, two wheels on one side of the target-vehicle are visible.
The heading angle $\varphi$ of the target-vehicle is determined by the slope of the $line(RW, FW)$, as shown in Fig. \ref{f_3.2.2}(a).
\begin{equation}\label{equ1}
    \varphi = arctan ( \frac{FW_y - RW_y}{FW_x - RW_x} )
\end{equation}
The positions of the left-front corner $A$ and the left-rear corner $B$ of the target-vehicle as well as their midpoint $M$ are determined with the help of its front overhang $fo$ and rear overhang $ro$, which are bounded with its type.
\begin{equation}
    \left (
    \begin{array}{lr}
        A_x \\
        A_y 
    \end{array}
    \right )
    =
    \left (
    \begin{array}{lr}
        FW_x + fo * cos(\varphi) \\
        FW_y + fo * sin(\varphi)
    \end{array}
    \right )
\end{equation}

\begin{equation}
    \left (
    \begin{array}{lr}
        B_x \\
        B_y 
    \end{array}
    \right )
    =
    \left (
    \begin{array}{lr}
        RW_x + ro * cos(\varphi) \\
        RW_y + ro * sin(\varphi)
    \end{array}
    \right )
    % B(x, y) = (RW_x + ro * cos(\varphi), RW_y + ro * sin(\varphi))
\end{equation}
\begin{equation}
    \left (
    \begin{array}{lr}
        M_x \\
        M_y 
    \end{array}
    \right )
    =
    \left (
    \begin{array}{lr}
        \frac{A_x + B_x}{2} \\
        \frac{A_y + B_y}{2}
    \end{array}
    \right )
    % M(x, y) = (\frac{A_x + B_x}{2}, \frac{A_y + B_y}{2})
\end{equation}
We acquire the position of center point $P$ by $line(M, P)$ and half of the width $w$ of the target-vehicle.
\begin{equation}
    \left (
    \begin{array}{lr}
        P_x \\
        P_y 
    \end{array}
    \right )
    =
    \left (
    \begin{array}{lr}
        M_x + \frac{w}{2} * cos(\varphi - 90) \\
        M_y + \frac{w}{2} * sin(\varphi - 90)
    \end{array}
    \right )
    % P(x, y) = (M_x + \frac{w}{2} * cos(\varphi - 90), M_y + \frac{w}{2} * sin(\varphi - 90))
\end{equation}
In the second case, one wheel and one bumper can be observed as shown in Fig. \ref{f_3.2.2}(b).
Similarly, the slope of the $line(RB, RW)$ determines the angle $\gamma$.
\begin{equation}
    \gamma = arctan ( \frac{RW_y - RB_y}{RW_x - RB_x} )
\end{equation}
The rear overhang $ro$ and half of the width $w$ of the target-vehicle $w$ indicate the angle $\phi$.
\begin{equation}
    \phi = arctan ( \frac{w/2}{ro} )
\end{equation}
We calculate the heading angle $\varphi$ of the target-vehicle based on the angle $\gamma$ and angle $\phi$.
\begin{equation}
    \varphi = \gamma - \phi
\end{equation}
The position of the center point $P$ can be determined by the position of the rear bumper $RB$ and half of the length $l$ of the target-vehicle.
\begin{equation}
    \left (
    \begin{array}{lr}
        P_x \\
        P_y 
    \end{array}
    \right )
    =
    \left (
    \begin{array}{lr}
        RB_x + l/2 *cos(\varphi) \\
        RB_y + l/2 *sin(\varphi)
    \end{array}
    \right )
    % P(x, y) = (RB_x + h/2 *cos(\varphi), RB_y + h/2 *sin(\varphi))
\end{equation}
In the third case, only one bumper is available. So we take the output of the Heading Angle Regression module as the final heading angle of the target-vehicle.

After acquiring the heading angle $\varphi$ and the center point $P$ of the target-vehicle, we can calculate the positions of its four corners by the width $w$ and length $l$ and illustrate it by a rectangle on the BEV Vector Map.

\section{Experiments and Results}

In this section, we evaluate the proposed visual perception method on accuracy, robustness and real-time performance. 
\begin{table}[!t]
  \begin{center}
    \caption{Comparison results on synthetic panorama dataset}
    \label{tab:experiment1}
    \begin{tabular}{lllllllll}
    \hline
    \textbf{Models} & \textbf{2D-AP} & \textbf{3D-mAP} & \textbf{AOS} & \textbf{IoU} &\textbf{Dist. Err.}\\
    \hline
      
      \cite{experiment1} & 0.447 & 0.203  & 0.157 & 0.265 & 1.143      \\
\hline
\cite{2020arxivmonocular} & 0.472 & 0.301  & 0.419 & 0.36  & 0.883      \\
\hline
\textbf{Ours}     & \textbf{0.573} & \textbf{0.362}  & \textbf{0.856} & \textbf{0.647} & \textbf{0.535}    \\
\hline
    \end{tabular}
  \end{center}
\end{table}
\subsection{Experimental settings}
The 3D detection methods usually require 3D labeled data from the public 3D datasets for training, such as the KITTI dataset \cite{KITTI}.
%The 3D detection methods usually require 3D labeled data for training, which can be obtained from public 3D datasets such as the KITTI dataset \cite{KITTI}.
Since we decompose the 3D object detection task into sub-tasks, our method doesn't require 3D labeled data. Currently, there is no public dataset suitable for our setting, so we collect and label the training data manually. 
The training dataset contains about 140,000 images collected by a surround-view camera system with four fisheye cameras.
%Our system is deployed on the Qualcomm 820A hardware platform which is equipped with a Adreno 530 GPU and a Hexagon 680 DSP. 
Our system is deployed on the Qualcomm 820A platform with an Adreno 530 GPU and a Hexagon 680 DSP. It can deliver about 1.2 TOPs (trillion operations per second). In comparison, the NVIDIA Xavier is for 30 TOPs, and Tesla's V3 ``Full Self-Driving" Computer is claimed to deliver 144 TOPs. There is no doubt that our system can operate on a platform with rather low computing ability.

\subsection{Results of the synthetic panorama dataset}

\begin{figure}[!t]
    \centering
    \includegraphics[height =4.6cm, width=8cm]{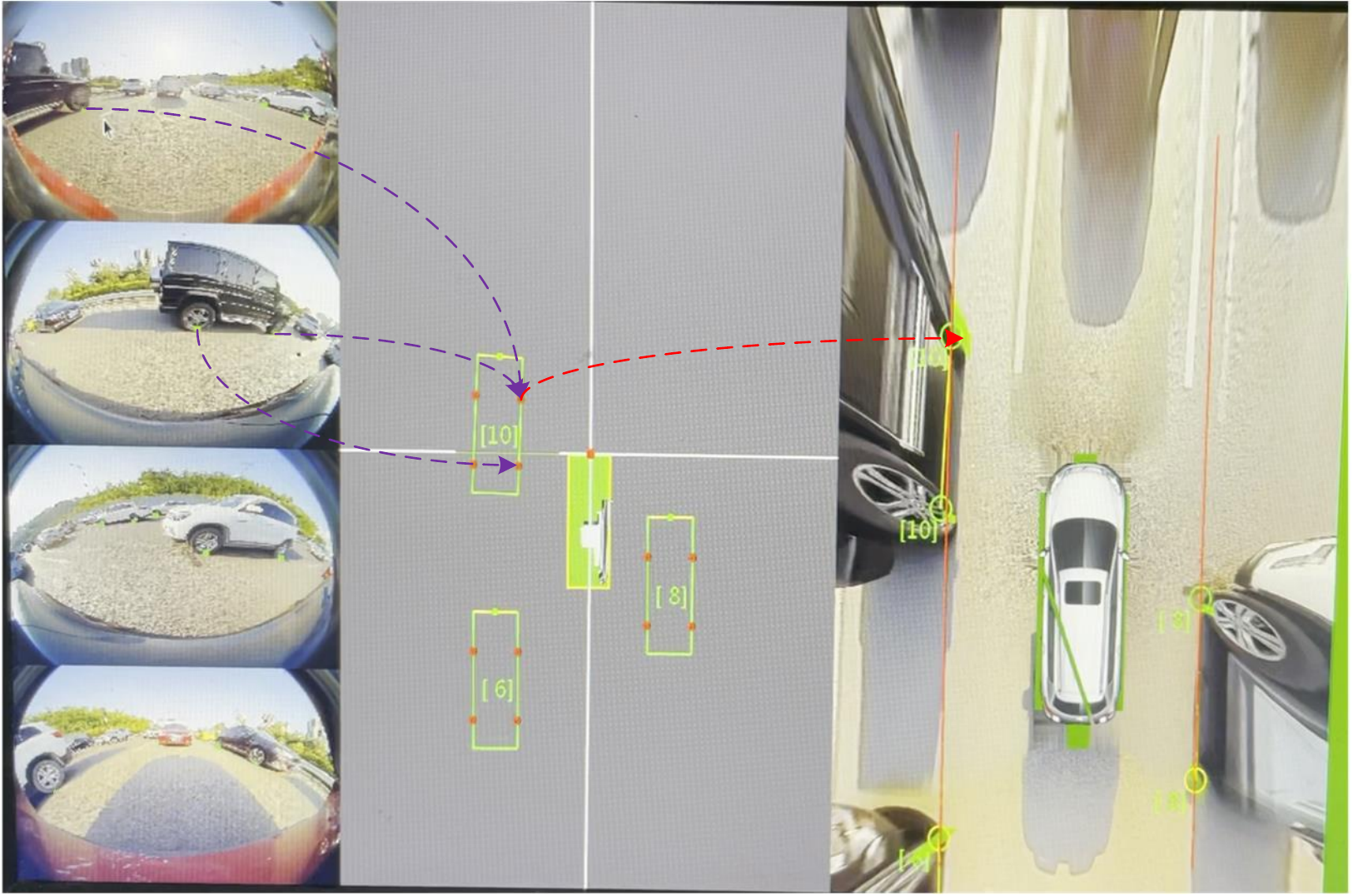}
    \caption{Visualization results of our perception system. Best view in color and zoom in.}
    \label{bev1}
\end{figure}

Following \cite{2020arxivmonocular}, we evaluate our results on the synthetic panorama dataset from \cite{experiment1}.
\cite{2020arxivmonocular} trained their 3D object detector on the KITTI 3D Object training set which is different from ours. The 2D and 3D metrics include 2D-AP (defined using IoU$>0.5$ between 2D bounding boxes following \cite{2020arxivmonocular}), 3D-mAP, average orientation similarity (AOS), mean 3D volumetric IoU, and mean Euclidean distance error. Table \ref{tab:experiment1} presents these results. 
Our method has achieved the best performance, especially on AOS and IoU metrics, which proves that our method is effective in decomposing the total task into sub-tasks, even without 3D labeled data. Fig. \ref{bev1} shows the visualization of the perception output.

%The output visualization of our system is shown in 8

\subsection{System positioning error analysis}
\begin{figure}[!t]
\centering
\
\subfloat[]{\includegraphics[width=4.2cm]{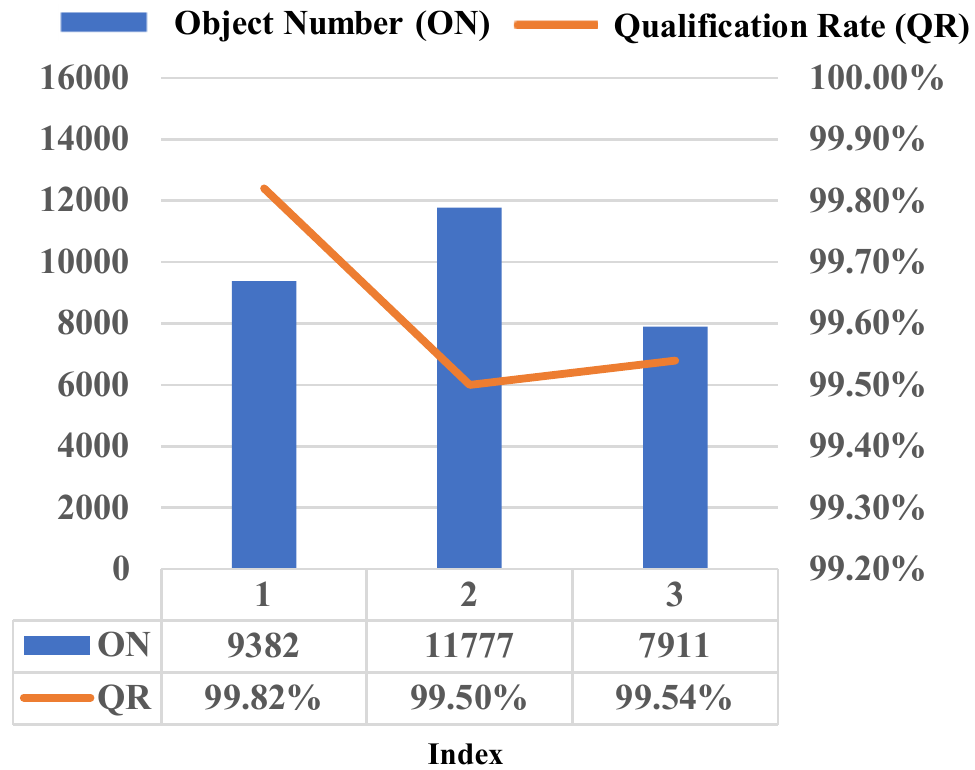}}
\subfloat[]{\includegraphics[width=4.3cm]{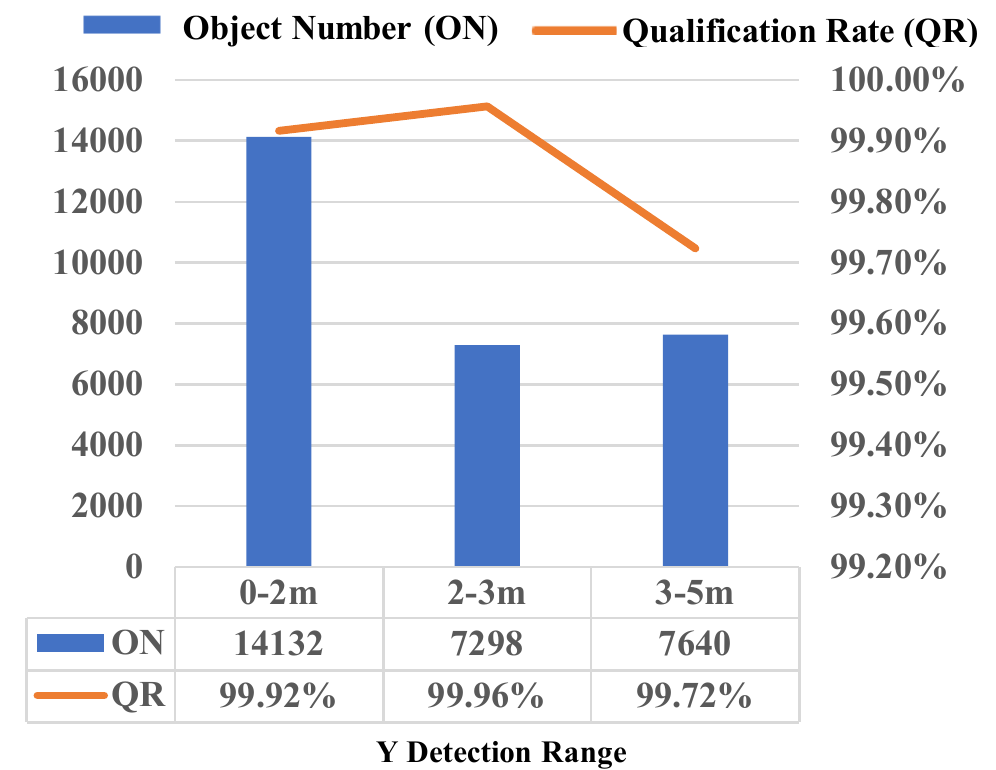}}

\caption{Evaluation of the system positioning qualification rate. Best view in color and zoom in.
%(a) Evaluation in three experiments along x direction. (b) Evaluation in three intervals along y direction.
}
%(a) The object number and qualification rate of object %position in three experiments along x direction. (b) The %object number and qualification rate of object position in %different intervals along y direction.}
\label{fig_exp01}
\end{figure}

We conduct three experiments to evaluate the object position accuracy. In these experiments, the numbers of objects perceived by our system are 9382, 11777, and 7911, respectively. In the evaluation of positioning error along the $x$-direction (Fig. \ref{f_3.2.2}), it is eligible when errors are smaller than 25 cm. As shown in Fig. \ref{fig_exp01}(a), the qualification rates of these three tests are 99.82$\%$, 99.50$\%$, and 99.54$\%$. 

As for $y$-direction (Fig. \ref{f_3.2.2}), since the strong relation between the positioning error and the distance from the target vehicle to the ego-vehicle, we divide the statistical errors into multiple intervals. In the experiments, the distance range is divided into 3 intervals of 0-2 meters, 2-3 meters, and 3-5 meters, while the requirements are no larger than 20 cm, 40 cm, and 50 cm. As shown in Fig. \ref{fig_exp01}(b), the qualification rates in the $y$-direction are 99.92$\%$, 99.96$\%$, and 99.72$\%$. These experiments effectively prove that our system has a high positioning accuracy.

\begin{figure}[t]
\centering
\subfloat[]{\includegraphics[width=4.2cm]{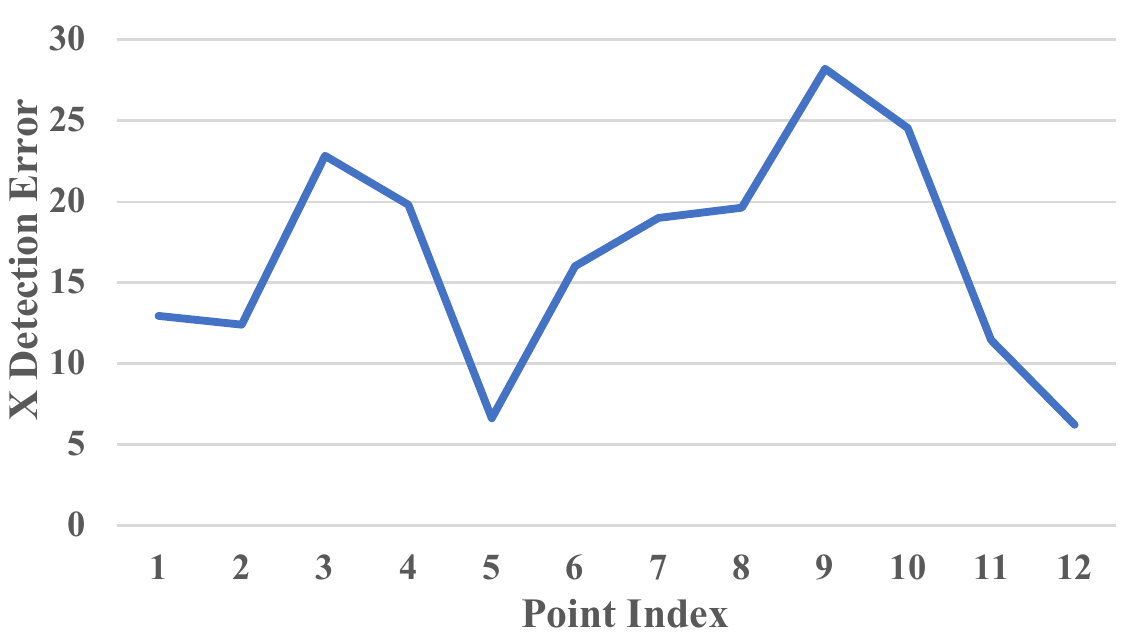}}
\  
\subfloat[]{\includegraphics[width=4.2cm]{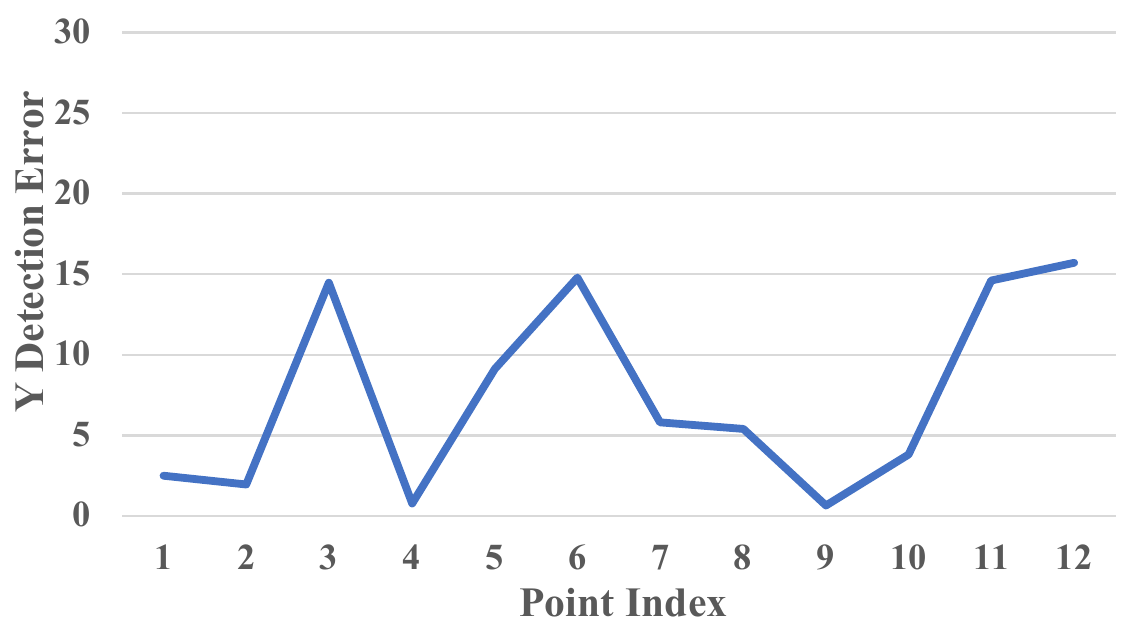}}

\caption{Evaluation of the positional accuracy on the ground with 5$\%$ slope gradient. Best view in color and zoom in.}
\label{fig_exp02}
\end{figure}

Since there is no ideal horizontal ground plane in practical application, we design an experiment to evaluate the positioning accuracy of our system on the ground with a 5$\%$ slope gradient. We randomly select 12 objects distributing from -2.5 to 2.5 meters along the $x$-direction and calculate their position errors. As shown in Fig. \ref{fig_exp02}(a), the result demonstrates that all the errors are less than 30 cm. Meanwhile, 12 points are randomly selected in the range of 1.5 to 3.5 meters along $y$-direction to analyze the position errors. As shown in Fig. \ref{fig_exp02}(b), the position errors of objects are less than 20 cm. It can be concluded that our system can work on the ground with a slight slope gradient, which proves the robustness of the proposed system in practical scenarios.
%=================================
%===========================================

\subsection{System delay analysis}
\begin{figure}[h]
    \centering
    \includegraphics[width=8.5cm]{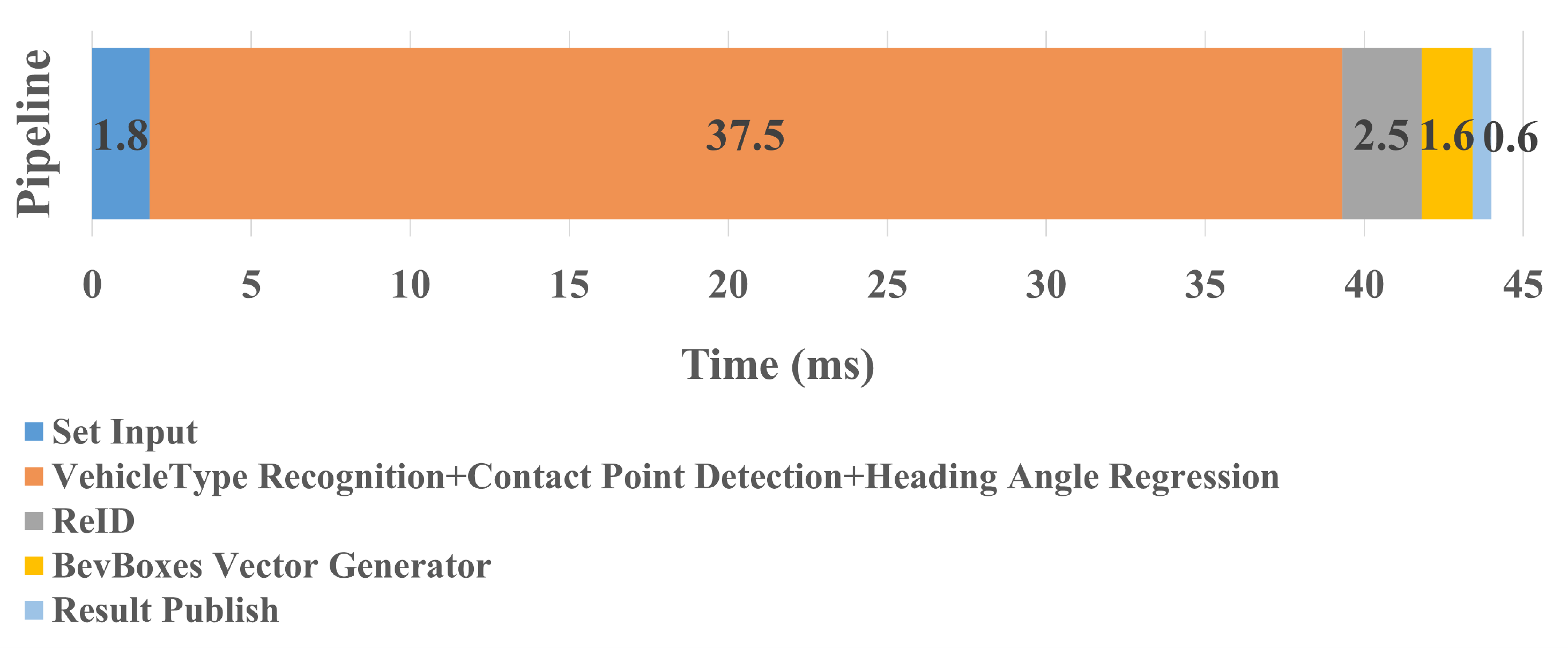}
    \caption{Time consumption tests on the hardware platform. The total time of our system's test is less than 45 ms, which can meet real-time requirements. Best view in color and zoom in.
    }
    \label{fig_exp03}
\end{figure}
We perform time consumption tests 
%for our system 
on the hardware platform, and Fig. \ref{fig_exp03} shows the latency time of each module.

The ``Set Input" and ``Result Publish" represent the input and output of the proposed system. 
``Vehicle Type Recognition", ``Contact Point Detection", and ``Heading Angle Regression" are introduced in the Method section.
The total time of our system's test is less than 45 ms. The ``Vehicle Type Recognition + Contact Point Detection + Heading Angle Regression" part takes the most time (37.5 ms) since it's the network inference module. The test is carried out on a serial setting and the total time is the sum of the running time for four input pictures in series.

In real engineering applications, the parallel structure achieves a total time consumption of less than 40 ms, which can meet real-time requirements.

\section{CONCLUSIONS}

In this paper, we present an accurate and low-latency 3D vehicle perception solution with the surround-view fisheye camera system for autonomous driving. We are the first to propose a method for overcoming and avoiding the acquisition difficulty of the large-scale accurate 3D labeled truth data, by breaking down the 3D object detection task into some sub-tasks and obtain sufficient precision for mass production. The above solution can achieve a sub-120 ms latency from view to multidimensional vector output, with average errors less than 0.25 meters at a distance of 5 meters. Furthermore, the associated fisheye dataset will be public to facilitate the progress of relevant research in this field. In future studies, we will continue to optimize the performance of the 3D perception system.

    %%%%%%%%%%%%%%%%%%%%%%%%%%%%%%%%%%%%%%%%%%%%%%%%%%%%%%%%%%%%%%%%%%%%%%%%%%%%%%%%
    % \addtolength{\textheight}{-12cm}
    %%%%%%%%%%%%%%%%%%%%%%%%%%%%%%%%%%%%%%%%%%%%%%%%%%%%%%%%%%%%%%%%%%%%%%%%%%%%%%%%
    %%%%%%%%%%%%%%%%%%%%%%%%%%%%%%%%%%%%%%%%%%%%%%%%%%%%%%%%%%%%%%%%%%%%%%%%%%%%%%%%
    \clearpage
    % \small
    \bibliographystyle{IEEEtran}
    \bibliography{refs}

\begin{thebibliography}{10}
\providecommand{\url}[1]{#1}
\csname url@samestyle\endcsname
\providecommand{\newblock}{\relax}
\providecommand{\bibinfo}[2]{#2}
\providecommand{\BIBentrySTDinterwordspacing}{\spaceskip=0pt\relax}
\providecommand{\BIBentryALTinterwordstretchfactor}{4}
\providecommand{\BIBentryALTinterwordspacing}{\spaceskip=\fontdimen2\font plus
\BIBentryALTinterwordstretchfactor\fontdimen3\font minus
  \fontdimen4\font\relax}
\providecommand{\BIBforeignlanguage}[2]{{%
\expandafter\ifx\csname l@#1\endcsname\relax
\typeout{** WARNING: IEEEtran.bst: No hyphenation pattern has been}%
\typeout{** loaded for the language `#1'. Using the pattern for}%
\typeout{** the default language instead.}%
\else
\language=\csname l@#1\endcsname
\fi
#2}}
\providecommand{\BIBdecl}{\relax}
\BIBdecl

\bibitem{2020IROSVision}
J.~Monica and M.~Campbell, ``Vision only 3-d shape estimation for autonomous
  driving,'' in \emph{{IEEE/RSJ} International Conference on Intelligent Robots
  and Systems, {IROS} 2020}.\hskip 1em plus 0.5em minus 0.4em\relax {IEEE},
  2020, pp. 1676--1683.

\bibitem{2020IROSAccurate}
K.~Strobel, S.~Zhu, R.~Chang, and S.~Koppula, ``Accurate, low-latency visual
  perception for autonomous racing: Challenges,mechanisms, and practical
  solutions,'' in \emph{{IEEE/RSJ} International Conference on Intelligent
  Robots and Systems, {IROS} 2020}.\hskip 1em plus 0.5em minus 0.4em\relax
  IEEE, 2020, pp. 1969--1975.

\bibitem{2016arxivvehicle}
B.~Li, T.~Zhang, and T.~Xia, ``Vehicle detection from 3d lidar using fully
  convolutional network,'' \emph{arXiv preprint arXiv:1608.07916}, 2016.

\bibitem{2018IVreal}
I.~Baek, A.~Davies, G.~Yan, and R.~R. Rajkumar, ``Real-time detection,
  tracking, and classification of moving and stationary objects using multiple
  fisheye images,'' in \emph{2018 IEEE Intelligent Vehicles Symposium
  (IV)}.\hskip 1em plus 0.5em minus 0.4em\relax IEEE, 2018, pp. 447--452.

\bibitem{ng2020bev}
M.~H. Ng, K.~Radia, J.~Chen, D.~Wang, I.~Gog, and J.~E. Gonzalez, ``Bev-seg:
  Bird's eye view semantic segmentation using geometry and semantic point
  cloud,'' \emph{arXiv preprint arXiv:2006.11436}, 2020.

\bibitem{mallot1991inverse}
H.~A. Mallot, H.~H. B{\"u}lthoff, J.~Little, and S.~Bohrer, ``Inverse
  perspective mapping simplifies optical flow computation and obstacle
  detection,'' \emph{Biological Cybernetics}, vol.~64, no.~3, pp. 177--185,
  1991.

\bibitem{2019ICCVaccurate}
X.~Ma, Z.~Wang, H.~Li, P.~Zhang, W.~Ouyang, and X.~Fan, ``Accurate monocular 3d
  object detection via color-embedded 3d reconstruction for autonomous
  driving,'' in \emph{Proceedings of the IEEE/CVF International Conference on
  Computer Vision}, 2019, pp. 6851--6860.

\bibitem{2019AAAImonogrnet}
Z.~Qin, J.~Wang, and Y.~Lu, ``Monogrnet: A geometric reasoning network for
  monocular 3d object localization,'' in \emph{Proceedings of the AAAI
  Conference on Artificial Intelligence}, vol.~33, no.~01, 2019, pp.
  8851--8858.

\bibitem{2019IVdeep}
Y.~Kim and D.~Kum, ``Deep learning based vehicle position and orientation
  estimation via inverse perspective mapping image,'' in \emph{2019 IEEE
  Intelligent Vehicles Symposium (IV)}.\hskip 1em plus 0.5em minus 0.4em\relax
  IEEE, 2019, pp. 317--323.

\bibitem{2019CVPRmulti}
M.~Liang, B.~Yang, Y.~Chen, R.~Hu, and R.~Urtasun, ``Multi-task multi-sensor
  fusion for 3d object detection,'' in \emph{Proceedings of the IEEE/CVF
  Conference on Computer Vision and Pattern Recognition}, 2019, pp. 7345--7353.

\bibitem{2020CVPRWsmoke}
Z.~Liu, Z.~Wu, and R.~T{\'o}th, ``Smoke: single-stage monocular 3d object
  detection via keypoint estimation,'' in \emph{Proceedings of the IEEE/CVF
  Conference on Computer Vision and Pattern Recognition Workshops}, 2020, pp.
  996--997.

\bibitem{2019ICCVwoodscape}
S.~Yogamani, C.~Hughes, J.~Horgan, G.~Sistu, P.~Varley, D.~O'Dea,
  M.~Uric{\'a}r, S.~Milz, M.~Simon, K.~Amende \emph{et~al.}, ``Woodscape: A
  multi-task, multi-camera fisheye dataset for autonomous driving,'' in
  \emph{Proceedings of the IEEE/CVF International Conference on Computer
  Vision}, 2019, pp. 9308--9318.

\bibitem{2016cvpryou}
J.~Redmon, S.~Divvala, R.~Girshick, and A.~Farhadi, ``You only look once:
  Unified, real-time object detection,'' in \emph{Proceedings of the IEEE
  Conference on Computer Vision and Pattern Recognition}, 2016, pp. 779--788.

\bibitem{headingangle}
A.~Mousavian, D.~Anguelov, J.~Flynn, and J.~Kosecka, ``3d bounding box
  estimation using deep learning and geometry,'' in \emph{Proceedings of the
  IEEE Conference on Computer Vision and Pattern Recognition}, 2017, pp.
  7074--7082.

\bibitem{headingangle1}
J.~Ku, A.~D. Pon, and S.~L. Waslander, ``Monocular 3d object detection
  leveraging accurate proposals and shape reconstruction,'' in
  \emph{Proceedings of the IEEE/CVF Conference on Computer Vision and Pattern
  Recognition}, 2019, pp. 11\,867--11\,876.

\bibitem{KITTI}
A.~Geiger, P.~Lenz, and R.~Urtasun, ``Are we ready for autonomous driving? the
  kitti vision benchmark suite,'' in \emph{2012 IEEE Conference on Computer
  Vision and Pattern Recognition}.\hskip 1em plus 0.5em minus 0.4em\relax IEEE,
  2012, pp. 3354--3361.

\bibitem{2019arxivfisheyemultinet}
P.~Maddu, W.~Doherty, G.~Sistu, I.~Leang, M.~Uricar, S.~Chennupati, H.~Rashed,
  J.~Horgan, C.~Hughes, and S.~Yogamani, ``Fisheyemultinet: Real-time
  multi-task learning architecture for surround-view automated parking
  system,'' \emph{arXiv preprint arXiv:1912.11066}, 2019.

\bibitem{2021IRALground}
Y.~Liu, Y.~Yuan, and M.~Liu, ``Ground-aware monocular 3d object detection for
  autonomous driving,'' \emph{IEEE Robotics and Automation Letters}, 2021.

\bibitem{2019arxivobjects}
X.~Zhou, D.~Wang, and P.~Kr{\"a}henb{\"u}hl, ``Objects as points,'' \emph{arXiv
  preprint arXiv:1904.07850}, 2019.

\bibitem{2018IROSend}
M.~Toromanoff, E.~Wirbel, F.~Wilhelm, C.~Vejarano, X.~Perrotton, and
  F.~Moutarde, ``End to end vehicle lateral control using a single fisheye
  camera,'' in \emph{2018 IEEE/RSJ International Conference on Intelligent
  Robots and Systems (IROS)}.\hskip 1em plus 0.5em minus 0.4em\relax IEEE,
  2018, pp. 3613--3619.

\bibitem{2014ICCPomnidirectional}
M.~Drulea, I.~Szakats, A.~Vatavu, and S.~Nedevschi, ``Omnidirectional stereo
  vision using fisheye lenses,'' in \emph{2014 IEEE 10th International
  Conference on Intelligent Computer Communication and Processing
  (ICCP)}.\hskip 1em plus 0.5em minus 0.4em\relax IEEE, 2014, pp. 251--258.

\bibitem{2020ICRAfisheyedistancenet}
V.~R. Kumar, S.~A. Hiremath, M.~Bach, S.~Milz, C.~Witt, C.~Pinard, S.~Yogamani,
  and P.~M{\"a}der, ``Fisheyedistancenet: Self-supervised scale-aware distance
  estimation using monocular fisheye camera for autonomous driving,'' in
  \emph{2020 IEEE International Conference on Robotics and Automation
  (ICRA)}.\hskip 1em plus 0.5em minus 0.4em\relax IEEE, 2020, pp. 574--581.

\bibitem{2015IV360}
M.~Bertozzi, L.~Castangia, S.~Cattani, A.~Prioletti, and P.~Versari, ``360
  detection and tracking algorithm of both pedestrian and vehicle using fisheye
  images,'' in \emph{2015 IEEE Intelligent Vehicles Symposium (IV)}.\hskip 1em
  plus 0.5em minus 0.4em\relax IEEE, 2015, pp. 132--137.

\bibitem{2018ECCVWsemantic}
G.~Blott, M.~Takami, and C.~Heipke, ``Semantic segmentation of fisheye
  images,'' in \emph{Proceedings of the European Conference on Computer Vision
  (ECCV) Workshops}, 2018, pp. 181--196.

\bibitem{2018CVPRWnear}
V.~R. Kumar, S.~Milz, C.~Witt, M.~Simon, K.~Amende, J.~Petzold, S.~Yogamani,
  and T.~Pech, ``Near-field depth estimation using monocular fisheye camera: A
  semi-supervised learning approach using sparse lidar data,'' in \emph{CVPR
  Workshop}, vol.~7, 2018.

\bibitem{2021WACVsyndistnet}
V.~R. Kumar, M.~Klingner, S.~Yogamani, S.~Milz, T.~Fingscheidt, and P.~Mader,
  ``Syndistnet: Self-supervised monocular fisheye camera distance estimation
  synergized with semantic segmentation for autonomous driving,'' in
  \emph{Proceedings of the IEEE/CVF Winter Conference on Applications of
  Computer Vision}, 2021, pp. 61--71.

\bibitem{2020accessfisheyedet}
T.~Li, G.~Tong, H.~Tang, B.~Li, and B.~Chen, ``Fisheyedet: A self-study and
  contour-based object detector in fisheye images,'' \emph{IEEE Access},
  vol.~8, pp. 71\,739--71\,751, 2020.

\bibitem{2019arxivfisheyemodnet}
M.~Yahiaoui, H.~Rashed, L.~Mariotti, G.~Sistu, I.~Clancy, L.~Yahiaoui, V.~R.
  Kumar, and S.~Yogamani, ``Fisheyemodnet: Moving object detection on
  surround-view cameras for autonomous driving,'' \emph{arXiv preprint
  arXiv:1908.11789}, 2019.

\bibitem{2020CVPRWvehicle}
Z.~Wu, M.~Wang, L.~Yin, W.~Sun, J.~Wang, and H.~Wu, ``Vehicle re-id for
  surround-view camera system,'' \emph{arXiv preprint arXiv:2006.16503}, 2020.

\bibitem{2019ICCVWrotinvmtl}
B.~Arsenali, P.~Viswanath, and J.~Novosel, ``Rotinvmtl: Rotation invariant
  multinet on fisheye images for autonomous driving applications,'' in
  \emph{Proceedings of the IEEE/CVF International Conference on Computer Vision
  Workshops}, 2019, pp. 2373--2382.

\bibitem{2020arxivmonocular}
E.~Plaut, E.~B. Yaacov, and B.~E. Shlomo, ``Monocular 3d object detection in
  cylindrical images from fisheye cameras,'' \emph{arXiv preprint
  arXiv:2003.03759}, 2020.

\bibitem{2019ICCVWdesoiling}
M.~Uric{\'a}r, J.~Ulicny, G.~Sistu, H.~Rashed, P.~Krizek, D.~Hurych,
  A.~Vobecky, and S.~Yogamani, ``Desoiling dataset: Restoring soiled areas on
  automotive fisheye cameras,'' in \emph{Proceedings of the IEEE/CVF
  International Conference on Computer Vision Workshops}, 2019, pp. 4273--4279.

\bibitem{2021arxivomnidet}
V.~R. Kumar, S.~Yogamani, H.~Rashed, G.~Sitsu, C.~Witt, I.~Leang, S.~Milz, and
  P.~M{\"a}der, ``Omnidet: Surround view cameras based multi-task visual
  perception network for autonomous driving,'' \emph{arXiv preprint
  arXiv:2102.07448}, 2021.

\bibitem{2021IET4net}
Y.~Wu, S.~Feng, X.~Huang, and Z.~Wu, ``L4net: An anchor-free generic object
  detector with attention mechanism for autonomous driving,'' \emph{IET
  Computer Vision}, 2021.

\bibitem{2016cvprmonocular}
X.~Chen, K.~Kundu, Z.~Zhang, H.~Ma, S.~Fidler, and R.~Urtasun, ``Monocular 3d
  object detection for autonomous driving,'' in \emph{Proceedings of the IEEE
  Conference on Computer Vision and Pattern Recognition}, 2016, pp. 2147--2156.

\bibitem{2017CVPRdeep}
F.~Chabot, M.~Chaouch, J.~Rabarisoa, C.~Teuliere, and T.~Chateau, ``Deep manta:
  A coarse-to-fine many-task network for joint 2d and 3d vehicle analysis from
  monocular image,'' in \emph{Proceedings of the IEEE Conference on Computer
  Vision and Pattern Recognition}, 2017, pp. 2040--2049.

\bibitem{2019ICCVm3d}
G.~Brazil and X.~Liu, ``M3d-rpn: Monocular 3d region proposal network for
  object detection,'' in \emph{Proceedings of the IEEE/CVF International
  Conference on Computer Vision}, 2019, pp. 9287--9296.

\bibitem{lecun1998gradient}
Y.~LeCun, L.~Bottou, Y.~Bengio, and P.~Haffner, ``Gradient-based learning
  applied to document recognition,'' \emph{Proceedings of the IEEE}, vol.~86,
  no.~11, pp. 2278--2324, 1998.

\bibitem{experiment1}
G.~P. de~La~Garanderie, A.~A. Abarghouei, and T.~P. Breckon, ``Eliminating the
  blind spot: Adapting 3d object detection and monocular depth estimation to
  360 panoramic imagery,'' in \emph{Proceedings of the European Conference on
  Computer Vision (ECCV)}, 2018, pp. 789--807.

\end{thebibliography}

    %%%%%%%%%%%%%%%%%%%%%%%%%%%%%%%%%%%%%%%%%%%%%%%%%%%%%%%%%%%%%%%%%%%%%%%%%%%%%%%%

\end{document}